%% file: 8-main.tex
\pdfoutput=1

\documentclass[11pt]{article}

\usepackage[preprint]{acl}

\usepackage{times}
\usepackage{latexsym}
\usepackage{booktabs}
\usepackage{multirow}
\usepackage{adjustbox}  

\usepackage{booktabs}   
\usepackage{multirow}   
\usepackage{tabularx}   
\usepackage[most]{tcolorbox} 
\usepackage{lipsum}          

\NewTColorBox{instructionbox}{O{red}O{Instruction:}}{%
  enhanced,
  breakable,
  colback=#1!5,
  colframe=#1!70!black,
  boxrule=1pt,
  left=8pt,right=8pt,top=8pt,bottom=8pt,
  sharp corners,
  title=#2,
  fonttitle=\bfseries\color{white},
  colbacktitle=#1!80!black,
  coltitle=white,
  titlerule=0pt,
  toptitle=4pt,
  bottomtitle=4pt,
}

\usepackage[T1]{fontenc}

\usepackage[utf8]{inputenc}

\usepackage{microtype}

\usepackage{inconsolata}

\usepackage{graphicx}
\usepackage{xspace}          
\newcommand{\model}{RouteLLM\xspace}
\usepackage{amsmath}
\usepackage{amssymb}
\usepackage{comment}

\usepackage{graphicx}
\usepackage{subcaption}
\usepackage{amsfonts}
\usepackage{bbm}

\title{Bridging Language and Navigation: A Multi-Agent LLM System for Flexible Route Recommendation}

\title{Constraint-Aware Route Recommendation from Natural Language \\ via Hierarchical LLM Agents}

\author{
\textbf{Tao Zhe}\textsuperscript{1}, 
\textbf{Rui Liu}\textsuperscript{1},
\textbf{Fateme Memar}\textsuperscript{1}, 
\textbf{Xiao Luo}\textsuperscript{2}, \\
\textbf{Wei Fan}\textsuperscript{3},  
\textbf{Xinyue Ye}\textsuperscript{4},  
\textbf{Zhongren Peng}\textsuperscript{5}, 
\textbf{Dongjie Wang}\textsuperscript{1\dag} 
\\
\textsuperscript{1}University of Kansas \quad 
\textsuperscript{2}UW--Madison \quad 
\textsuperscript{3}University of Auckland \\
\textsuperscript{4}University of Alabama \quad 
\textsuperscript{5}University of Florida 
\\
\small{\{taozhe, rayliu, amemar, wangdongjie\}@ku.edu} \\
\small{xiao.luo@wisc.edu, wei.fan@auckland.ac.nz, xye10@ua.edu, zpeng@dcp.ufl.edu} \\
\small{\textsuperscript{\dag}Corresponding author}
}

\begin{document}
\maketitle

\input{0-Abstract}

\input{1-Intro}

\input{2-Related_work}


\input{4-Methodogy}

\input{5-Experiment}
\input{6-Conclusion}

\section*{Limitations}
\textbf{Simulated Dataset and Lack of Real-world Data.}  
To our knowledge, no comparable benchmarks currently exist. While our simulated grid-based dataset can approximate real-world scenarios and the case study illustrates effectiveness, more complex and systematic real-world datasets would enable a more complete framework and more robust evaluation. Route recommendation in real-world contexts is inherently complex, and unified datasets are difficult to obtain due to diverse objectives and incomplete information. In future work, we aim to construct real-world datasets to address this gap.  

\noindent \textbf{Enhancements to the framework.}  
Several components of our framework offer clear avenues for future improvement. For instance, the Management Agent could be extended to handle more complex contexts and constraints, such as temporal dependencies, and the multi-agent architecture itself warrants further refinement. Promising future directions also include supporting multimodal queries and outputs, and integrating reinforcement learning (RL) for enhanced personalization and route optimization. Beyond the core framework, \model can be flexibly integrated with external capabilities, such as online Point-of-Interest (POI) recommendation \cite{wang2022online} and advanced spatial-temporal models \cite{wang2021towards,zhou2020deep,wang2018deepstcl,chen2021integrated}, to provide more diversified and effective routing functions.

\noindent \textbf{Demonstration focus rather than strict comparison.}  
Given the limitations of the current dataset, conducting strict, large-scale comparative analyses is challenging. Our experiments, therefore, emphasize a qualitative demonstration of \model's core capabilities rather than exhaustive performance benchmarking. We are also aware of several promising avenues for future improvement. These include enhancing the capabilities of the planning stage, improving the efficiency of large-scale tool invocation, incorporating online learning to better adapt to individual user preferences, and supporting multi-modal inputs. Advancing research in these directions will likely require the construction of more comprehensive, real-world datasets.


\bibliography{TZ}

\appendix

\input{7-Appendix}

\end{document}

%% file: 0-Abstract.tex
\begin{abstract}
Route recommendation aims to provide users with optimal travel plans that satisfy diverse and complex requirements.
Classical routing algorithms (e.g., shortest-path and constraint-aware search) are efficient but assume structured inputs and fixed objectives, limiting adaptability to natural-language queries. 
Recent LLM-based approaches enhance flexibility but struggle with spatial reasoning and the joint modeling of route-level and POI-level preferences.
To address these limitations, we propose \model, a hierarchical multi-agent framework that grounds natural-language intents into constraint-aware routes.
It first parses user queries into structured intents including POIs, paths, and constraints. A manager agent then coordinates specialized sub-agents: a constraint agent that resolves and formally check constraints, a POI agent that retrieves and ranks candidate POIs, and a path refinement agent that refines routes via a routing engine with preference-conditioned costs. 
A final verifier agent ensures constraint satisfaction and produces the final route with an interpretable rationale.
This design bridges linguistic flexibility and spatial structure, enabling reasoning over route feasibility and user preferences.
Experiments show that our method reliably grounds textual preferences into constraint-aware routes, improving route quality and preference satisfaction over classical methods.
Our code and data are publicly available\footnote{\url{https://anonymous.4open.science/r/RouteLLM_ACL-DE52}}.    
\end{abstract}

%% file: 1-Intro.tex
\section{Introduction}

\begin{figure}[t]
  \includegraphics[width=\columnwidth]{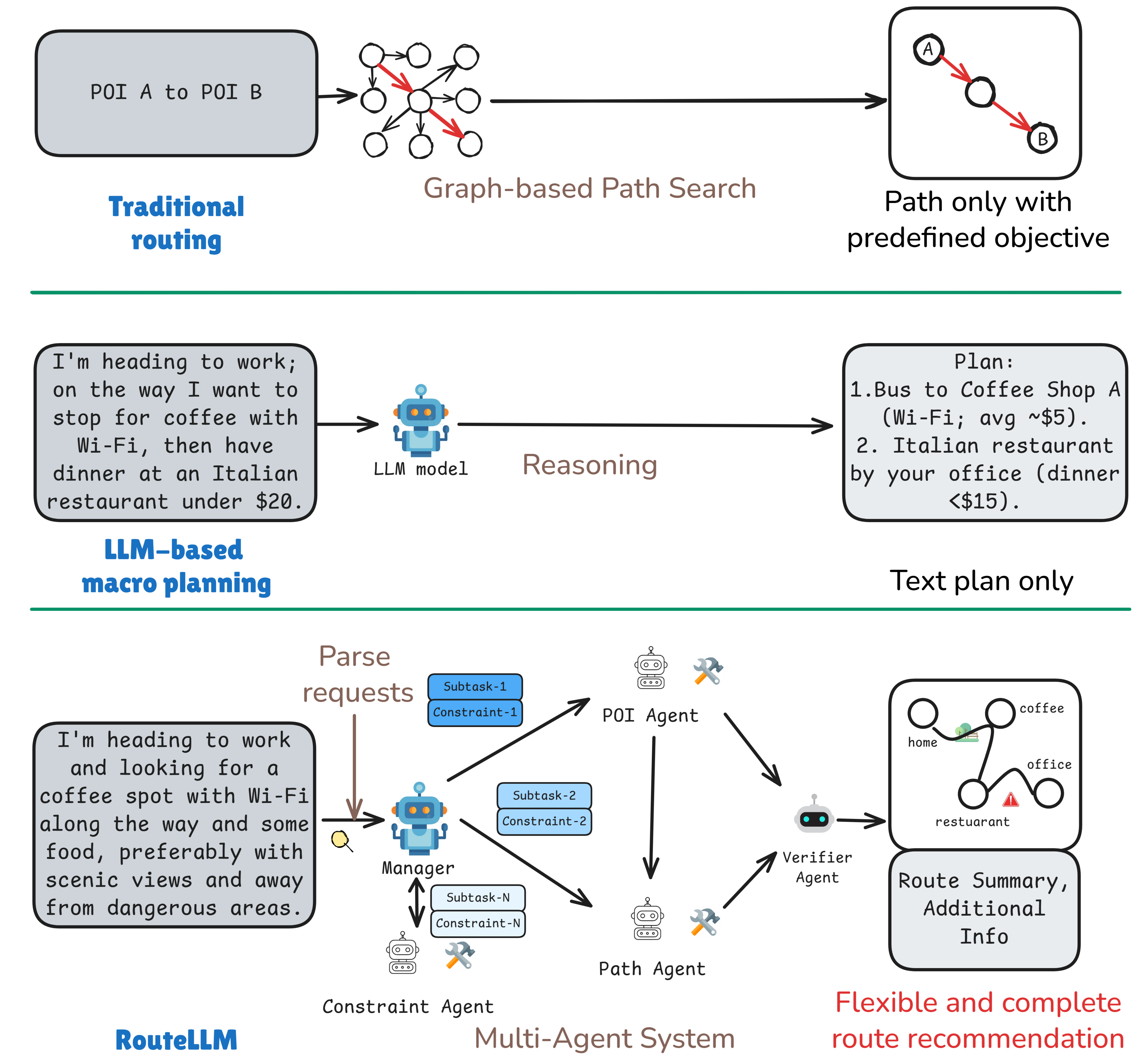}
  \caption{\model introduces a hierarchical multi-agent framework that bridges LLM reasoning with classical routing, enabling flexible, interpretable, and constraint-aware route recommendation.}
  \label{fig:experiments}
  \vspace{-0.5cm}
\end{figure}



Route recommendation approaches\citep{zhang2024survey} have become indispensable in modern daily life, spanning from high-level trip planning to detailed turn-by-turn navigation across diverse transportation scenarios. The rapid advancement of information technologies—including electronic devices, data storage, and 5G networks—has enabled the collection of vast spatiotemporal trajectory data reflecting daily commutes, travels, and activities \citep{zheng2009mining}. 
Simultaneously, as people navigate a complex landscape of travel options, the growing demand for intelligent, personalized route recommendations has made the design of such systems a task of significant research and practical value, with applications spanning navigation, logistics, and comprehensive travel planning\citep{zheng2015trajectory,liu2020integrating,xie2024travelplanner}. 

Traditional route recommendation approaches have established a solid 
foundation for path-finding through various algorithmic paradigms. Classical search-based algorithms \citep{kleinberg2006tardos} and heuristic search methods \citep{hart1968formal,mandow2008multiobjective,ulloa2020simple,zhang2022pex,zhang2024aa} have demonstrated effectiveness in identifying optimal paths based on predefined objectives such as shortest distance or minimal travel time. With advances in data storage and computing technologies, deep learning methods \citep{jain2021neuromlr,almasan2022deep,kong2024rpmtd,peng2021urban} have further enhanced route optimization by learning complex patterns from large-scale trajectory and network data. 
These approaches excel at providing precise, computationally efficient solutions for well-defined routing problems on graph structures. However, traditional methods face fundamental limitations in addressing modern users' diverse and evolving requirements. Most approaches operate within rigid frameworks optimizing for fixed objectives, lacking the flexibility to accommodate personalized preferences, contextual constraints, or complex multi-criteria decision-making. Furthermore, these systems require rigid, predefined parameter settings, making them inherently unsuitable for handling natural language queries or adapting to nuanced routing needs beyond standard optimization metrics.


Meanwhile, LLMs offer promising natural language understanding and reasoning capabilities for route recommendation, but current approaches reveal fundamental limitations across different paradigms.
One category of methods attempts to directly input raw map data to LLMs for end-to-end route planning\citep{zhang2024llm4dyg,chen2024can,aghzal2023can,li2024urbangpt}. These approaches prove highly ineffective due to LLMs' inherently poor spatial understanding and inability to efficiently process large-scale geographical information, often resulting in spatially inconsistent or infeasible route suggestions. Another line of work focuses on enhancing traditional path search algorithms with LLM-guided optimization\citep{xiao2023llm,gupta-li-2024-training,meng2024llm,xiao2025llm}, but these methods remain constrained to fixed optimization objectives and fail to leverage LLMs' flexible reasoning strength for flexible user interaction. A third category treats route recommendation as a high-level planning problem, employing LLMs for text-based reasoning and itinerary generation\citep{xie2024travelplanner,zhang2024planning,wong2023autonomous}. While these approaches successfully handle macro-level trip planning, they lack the spatial reasoning capabilities necessary for detailed route optimization and fail to bridge the gap between high-level planning and precise navigation. Consequently, existing LLM-based methods address only partial aspects of the route recommendation challenge, leaving a critical gap for systems that can seamlessly integrate objective flexibility with spatial precision.

To address the limitations of existing approaches, we propose \model, a hierarchical multi-agent framework that synergistically combines the precise path-finding of traditional algorithms with the flexible reasoning of LLM agents. Inspired by human planning, \model decomposes complex, multi-objective natural language requests into manageable sub-tasks with corresponding constraints. The framework employs specialized agents for parsing requests, coordinating the execution of sub-tasks, and synthesizing the results into a comprehensive route solution.
This multi-agent design effectively addresses the core limitations of prior approaches by leveraging the complementary strengths of both technologies. By breaking down complex problems, the framework mitigates the LLM's poor spatial reasoning while maintaining the natural language flexibility users demand. The integration of traditional path-finding algorithms ensures spatial accuracy and efficiency, bridging the critical gap between natural language accessibility and spatial precision to provide a robust foundation for personalized route recommendation.

%% file: 2-Related_work.tex
\section{Related Work}

\subsection{Route Recommendation}
Early route recommendation research was dominated by classical path-finding algorithms. Foundational methods such as graph search \cite{kleinberg2006algorithm} and heuristic search \citep{hart1968formal,mandow2008multiobjective,ulloa2020simple,zhang2022pex,zhang2024aa} provided efficient solutions under single or multiple objectives, typically optimizing metrics like distance or travel time.  
With advances in data storage and computation, learning-based approaches emerged. Deep learning methods \citep{jain2021neuromlr,almasan2022deep,kong2024rpmtd,peng2021urban} leverage large-scale trajectory and spatiotemporal data for more sophisticated optimization in logistics, delivery, and urban planning. Despite their efficiency, both classical and deep-learning methods remain rigid, offering limited support for personalized preferences or multi-criteria decision-making.

\vspace{-2pt}

\subsection{LLMs for Spatial and Route Reasoning}
Large language models (LLMs) have been explored for their ability to interpret natural language queries and support flexible planning. Existing work falls into three categories, each with notable limitations:  
\textit{End-to-End LLM Planners} attempt to process raw map data directly \citep{zhang2024llm4dyg,chen2024can,aghzal2023can,li2024urbangpt}, but suffer from poor spatial comprehension and scalability issues, often producing infeasible routes.  
\textit{LLM-Augmented Search} enhances traditional heuristics with LLM guidance \citep{xiao2023llm,gupta2024training,meng2024llm,xiao2025llm}, improving efficiency but remaining bound to fixed objectives and failing to exploit LLMs’ full reasoning capabilities.  
\textit{High-Level Trip Planning} applies LLMs to itinerary generation \citep{xie2024travelplanner,zhang2024planning,wong2023autonomous}, enabling macro-level reasoning but lacking detailed spatial optimization.  
Bridging the gap between flexible natural language understanding and precise route optimization remains a central challenge. 

%% file: 4-Methodogy.tex
\section{Methodology}

\begin{figure*}[t]      
  \centering
  \includegraphics[width=\textwidth]{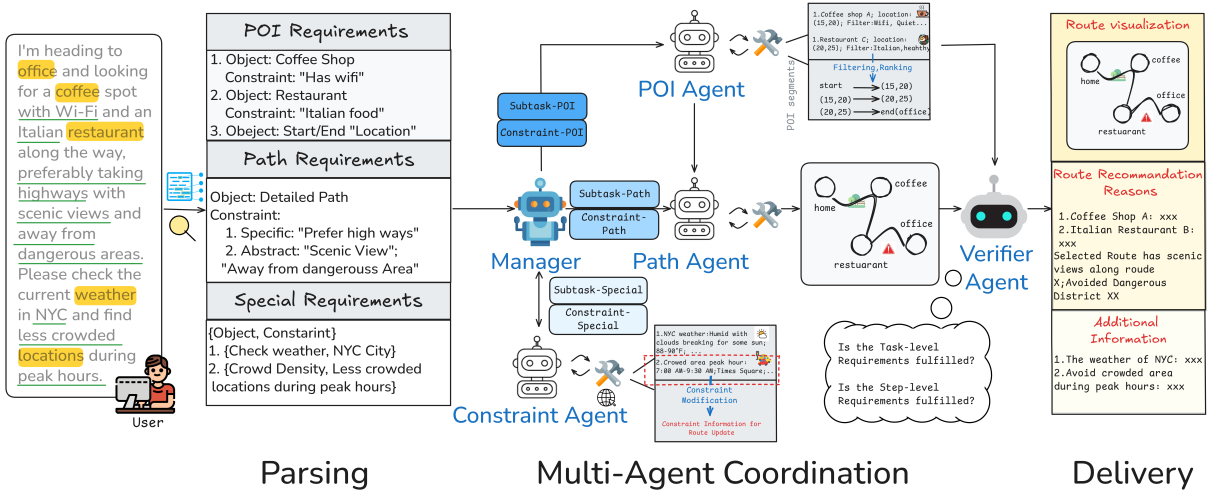}  
    \caption{The operational workflow of the \model framework. A user's natural language request is first parsed into structured requirements. Then, a manager agent coordinates specialized agents (POI, Path, Constraint Agent) to resolve subtasks. Finally, the system delivers an optimized route with detailed reasoning.}
  \label{fig:RouteLLM Framework}
\end{figure*}

To bridge the critical gap between flexible, nuanced user language and the rigid, mathematical precision of route optimization algorithms, we designed \model, a multi-agent framework as illustrated in Figure \ref{fig:RouteLLM Framework}. 
Our methodology systematically translates unstructured natural language into computable objectives through three key stages: (1) structured request parsing, (2) dependency-aware multi-agent coordination, and (3) adaptive, preference-weighted path planning.

\subsection{Parsing and Decomposing User Intent into a Computable Route Plan}
Human travel requests often blend multiple destinations with subjective preferences, such as scenery or specific amenities. Traditional route planners cannot interpret these nuanced requests because they rely on rigid A-to-B inputs.
The first stage of our framework bridges this gap by parsing unstructured requests into computable objectives and constraints. This converts natural user language into structured data for optimization algorithms.
To achieve this, our parsing approach decomposes requests using an [Object + Constraint] paradigm. Formally, the overall route planning task $\mathcal{T}$ is decomposed into interconnected sub-tasks:
$\mathcal{T} = \{ T_i \mid i \in S(K) = \{1, \ldots, K\} \}$ with corresponding constraints $\mathcal{C} = \{c_i(T_i) \leq 0 \mid i = 1,\dots,K \}$. 
Crucially, this process can easily preserve the full spectrum of user intent, from explicit requirements to nuanced preferences, enabling downstream components to plan personalized and efficient routes.

Specifically, the Parser Agent extracts three primary categories of routing requirements:
\underline{POI Requirements ($\mathcal{T}_{POI}$):} Location-specific requests with filtering constraints (e.g., "visit a coffee shop with WiFi and quiet environment, then an Italian restaurant"). These requirements specify venue types with attribute-based filters, formalized as tuples $(type_i, \{attr_{i,j}\})$ where $type_i$ denotes the venue category and $\{attr_{i,j}\}$ represents the set of required attributes.
\underline{Path Requirements ($\mathcal{T}_{path}$):} Route-level preferences comprising two distinct types:
\textbf{Specific constraints}: Well-defined binary restrictions (e.g., avoid toll roads, prefer highways) that can be directly encoded as hard constraints in the path search algorithm. \textbf{Abstract preferences:} Subjective requirements (e.g., scenic routes, energy efficiency, safety) requiring LLM-based interpretation to generate appropriate request-specific heuristic objectives as described in Section 3.3.
\underline{Special Requirements ($\mathcal{T}_{special}$):} Dynamic requirements that depend on real-time context or user-specific conditions not captured in the system's predefined ontology. Examples include weather-adaptive routing, live traffic event avoidance, and personalized accessibility accommodations. These requirements trigger auxiliary data retrieval and either (i) dynamically adjust constraints in both POI selection and path planning phases, or (ii) provide supplementary information without modification.

Among the three requirement categories, POI requirements translate directly into database queries, and path/special requirements can be encoded as hard constraints. Yet the core challenge lies in quantifying subjective Abstract preferences, which cannot be directly formalized.
To solve this, the Parser Agent uses contextual prompting to extract preference weights. 
Rather than relying on supervised sentiment analysis, which requires extensive labeled datasets and may suffer from generalization issues, our approach adapts chain-of-thought sentiment reasoning\cite{fei2023reasoning} for preference extraction to:
(1) identify abstract features mentioned in the input, (2) contextualize features with domain-specific knowledge, and (3) quantify preference intensity using a discrete scale.
We formalize the extracted preferences as a vector $\mathbf{p} = (p_1, p_2, \ldots, p_m) \in \{0, 0.5, 1\}^m$, where $m$ is the total number of possible abstract features (e.g., scenic, safe, energy-efficient), and each $p_i$ represents the user's preference intensity for feature $i$. The discrete scale captures three levels of user intent: $p_i = 0$ indicates the feature was not mentioned and can be ignored; $p_i = 0.5$ reflects a moderate preference where compromise is acceptable; and $p_i = 1$ represents a strong requirement treated as a hard constraint. This preference vector directly feeds into the multi-objective path planning algorithm (Section 3.3), where it guides the active objective decision and balance between competing objectives during route optimization.

\subsection{Mitigating LLM Spatial Weaknesses via Multi-Agent Coordination}
With the user's request structured into computable tasks, the next critical challenge is execution. Attempting to solve the entire plan with a single, monolithic LLM is ill-suited for this challenge for two primary reasons. 
First, LLMs exhibit a fundamental weakness in complex spatial reasoning, the very core of route planning. This deficit is often exacerbated by limited context windows, which can lead to forgotten constraints during a multi-step journey\cite{liu2023lost}.
Second, the problem itself is not uniform but is a composite of diverse sub-tasks. These range from structured database queries (for POIs) and algorithmic searches (for pathfinding) to dynamic requests requiring real-time context (like weather or traffic). Effective handling of such diversity requires a team of specialized agents, as a single, general-purpose LLM lacks both the specific tools and the contextual awareness needed for each distinct task.
Therefore, to mitigate these intrinsic model weaknesses and effectively manage the problem's heterogeneous nature, our framework employs a multi-agent paradigm. This approach decomposes the single, large challenge into manageable sub-problems that can be systematically coordinated and assigned to specialized agents.


\subsubsection{Hierarchical Planning and Decomposition}
The cornerstone of this multi-agent paradigm is hierarchical planning and decomposition, orchestrated by the Management Agent. This stage converts parsed tasks into an executable plan by addressing two key challenges. First, it manages logical dependencies in route planning (e.g., selecting destinations before planning paths). Second, it decomposes operations strategically to mitigate LLM spatial reasoning limitations.

The planning process operates on two tiers to ensure logical coherence and robust execution. 
At the task level, the Management Agent identifies dependencies between sub-tasks. For example, 
Road block conditions must precede path optimization because routes depend on passable roads. Based on these dependencies, the agent determines an appropriate execution sequence.
Beyond determining task order, the agent also decomposes individual tasks into finer-grained steps.
At the step level, each task is decomposed into specific actions. For instance, "find a route" is broken into steps like "identify waypoints," "calculate distances," and "optimize path." This decomposition mitigates LLM spatial reasoning limitations by providing focused context for each action.
Once the hierarchical plan is established, it is executed through a constraint-aware workflow. 
The Management Agent classifies constraints into two types: Local Constraints apply to individual sub-tasks (e.g., a POI's attributes), while Global Constraints span the entire journey (e.g., total travel budget). 
This classification informs the execution strategy. Tasks without interdependencies can be processed in parallel for efficiency. For example, querying weather and searching for POIs can run concurrently. However, tasks with dependencies or those that modify global constraints must be coordinated sequentially to ensure route validity. This two-tier approach balances parallel efficiency with dependency management for task-specific agents.

\subsubsection{Specialized Sub-Task Execution}
Following the decomposition and planning by the Management Agent, the framework enters the execution stage. Here, the individual sub-tasks are dispatched to a team of specialized agents, each equipped with domain-specific tools and contextual knowledge to handle the local constraints ($\mathcal{C}_l$) of their assigned tasks through focused expertise. 
Our framework employs three primary types of specialized agents, corresponding to the sub-task categories identified during parsing:


\paragraph{POI Selection and Filtering.}
For POI Requirements ($\mathcal{T}_{\text{POI}}$), the POI Selection Agent is invoked. Its role is to translate abstract location requests, such as "a quiet coffee shop with WiFi," into concrete, route-feasible geographical points.
To achieve this, the agent executes a three-step process: (1) it queries venue databases to retrieve candidates matching the required type (e.g., 'coffee shop'), (2) filters these results based on specific attribute constraints (e.g., 'has WiFi', 'quiet environment'), and (3) ranks the remaining venues using rating scores that incorporate user preferences.

\paragraph{Multi-Objective Path Optimization.}
For Path Requirements($\mathcal{T}_{path}$), the Path Optimization Agent is responsible for generating the actual travel segments $\mathbf{r}$ connecting the selected POIs.
The agent considers both user preferences (e.g., scenic routes, avoid highways) and objective constraints (e.g., road closures, traffic restrictions) to produce feasible and optimized routes.
We provide a more detailed treatment in Section 3.3.

\paragraph{Handling Dynamic and Special Constraints}
Finally, Special Requirements($\mathcal{T}_{special}$)
are managed by the Constraint Agent. This agent's crucial role is to ground the route plan in the real-world's dynamic context by processing requests that depend on live information. To handle the diverse nature of these requests, the agent operates in a flexible dual-mode design:
\textit{Information Mode:} For requests that simply require supplementary data (e.g., current weather conditions or local event schedules), the agent retrieves the relevant information and passes it as additional relevant information to the Verifier Agent for inclusion in the final summary. This mode enriches the user's plan without altering the route's constraints.
\textit{Constraint Modification Mode:} For requests that must actively alter the route (e.g., "avoid areas with current flooding"), the agent retrieves real-time data, translates the conditions into new routing constraints, and propagates these updates to other relevant agents.
This dual-mode architecture ensures that diverse special requirements are handled efficiently while maintaining a clear separation between the core routing logic and auxiliary information services.

\paragraph{Solution Synthesis and Presentation}
In the final stage of the framework, the verifier agent takes on the critical role of synthesizing the disparate outputs from all specialized agents into a single, cohesive, and user-friendly route recommendation. This process involves three key steps: integrating the partial results into coherent route plans, verifying these plans against global constraints, and finally, presenting the recommendations in an actionable, multi-modal format to users.
\underline{Route Integration and Alternative Generation:} The synthesis process begins by integrating the outputs from the specialized agents and creating complete plans. 

\underline{Global Constraint Verification:}
Once one or more complete candidate solutions ($\mathbf{s} = \{POI_{set}, \mathbf{r}\}$) are assembled, the Verifier Agent performs a crucial verification step against all global constraints ($\mathcal{C}_g$)
Unlike local constraints handled by individual agents, global constraints such as total travel time or cumulative budget require a holistic evaluation. The agent verifies each solution using the following function:
\begin{equation}
    \mathcal{C}_g(\mathbf{s}) = \begin{cases}
\text{feasible} & \text{if } g_j(\mathbf{s}) \leq \theta_j, \forall j \in \mathcal{G} \\
\text{infeasible} & \text{otherwise,}
\end{cases}
\end{equation}
where $g_j(\mathbf{s})$ is the value of the j-th global constraint, $\theta_j$ is its corresponding threshold and $\mathcal{G}$ is the set all global constraint. If a solution is infeasible, the agent applies a constraint relaxation strategy. Critical constraints (e.g., accessibility) may trigger a re-planning request to the Task-Specific Agents, while soft constraints (e.g., a slight budget overrun) can be relaxed based on the user's preference vector, with a notification included in the final output.
\underline{User-Oriented Presentation:}
After verifying the feasibility of the candidate routes, the final step is to transform the technical outputs into an intuitive and actionable format for the user. This is achieved through:
\textit{Natural Language Summarization:} Providing a concise summary of each route's key features (e.g., "This scenic route takes 15 minutes longer...") and detailed instructions.
\textit{Comparative Explanation:} Explicitly highlights differences between alternatives(e.g., "Option A is faster but misses the waterfront views...").
\textit{Visual Representation:} Generates interactive maps and structured itineraries.
This ensures users receive transparent, preference-aligned recommendations with clear justifications for routing decisions.·

\subsection{Adaptive Multi-Objective Path Planning}
Prior route recommendation methods rely on fixed, predefined objectives with rigid inputs. 
Our approach fundamentally reimagines this process by using LLMs as adaptive reasoning layers that translate human intentions into planning objectives, creating a dynamic bridge between natural language preferences and algorithmic optimization.
We model the road network as a graph $G=(V,E)$.
Each edge $e\in E$ is associated with $m$-dimensional non-negative cost vector $\mathbf{c}(e)$,
representing objectives such as distance and scenic cost. For a given path $\pi=\langle v_0,\ldots,v_k\rangle$, its cumulative cost is the vector sum of the costs of its edges:
\begingroup
\setlength{\abovedisplayskip}{6pt}
\setlength{\belowdisplayskip}{6pt}
\setlength{\abovedisplayshortskip}{0pt}
\setlength{\belowdisplayshortskip}{0pt}
\begin{equation}
\label{eq:path-sum}
\mathbf{g}(\pi)=\sum_{i=0}^{k-1}\mathbf{c}\!\left(v_i,v_{i+1}\right).   
\end{equation}
\endgroup
The LLM translates the user's abstract preferences into two key outputs:
an active objective set $\mathcal{A} \subseteq \{1, \dots, m\}$ and a corresponding weight vector $w$. 
This process tailors the search to the user's specific needs. For instance, an objective the user does not care about (e.g., 'scenic') is excluded from $\mathcal{A}$. A moderately important preference results in a reduced weight(e.g., $w_j = 0.5$) to the original cost landscape. 
while in the absence of specific preferences, the system defaults to a single-objective search($\mathcal{A} = {distance}$ using single-objective A*).
With the search parameters thus defined by the LLM, the subsequent challenge is to find the optimal set of routes that balance the user's preferences. Since there is typically no single "best" route in a multi-objective context, the goal is to find an $\epsilon$-approximate Pareto set using a suitable multi-objective search algorithm.(e.g., NAMOA*). This approach relies on the concept of $\epsilon$-dominance to compare solutions. For any two solutions with cost vectors $x$ and $y$, we say $x$ $\epsilon$-dominates $y$ if:
\begingroup
\setlength{\abovedisplayskip}{6pt}
\setlength{\belowdisplayskip}{6pt}
\setlength{\abovedisplayshortskip}{0pt}
\setlength{\belowdisplayshortskip}{0pt}
\begin{equation}
\begin{split}
    x \preceq_{\varepsilon} y \Leftrightarrow & \forall j \in \mathcal{A} : x_j \le (1 + \varepsilon)y_j \\
    & \quad\text{and}\quad \exists k \in \mathcal{A} : x_k < (1 + \varepsilon)y_k.
\end{split} 
\end{equation}
\endgroup
This process yields the top $k$ non-dominated solutions, which are then passed to the Delivery Agent. The agent provides a final natural language explanation of each route's unique trade-offs (e.g., "this route is faster but bypasses the scenic waterfront").

%% file: 5-Experiment.tex
\section{Experiment}

We evaluate \model through experiments highlighting three aspects: (1) parsing natural language into structured tasks, (2) adapting to user preference changes while respecting hard constraints, and (3) transfer to a real-world case study. To support this, we construct a benchmark dataset that mirrors realistic routing scenarios and enables systematic testing across varying complexity levels.    

\subsection{Experimental Setup}
A significant challenge in this domain is the lack of a suitable benchmark for language-driven route planning. To address this, our experimental setup consists of two main parts. First, for systematic and controlled testing, we constructed a new benchmark inspired by \cite{xie2024travelplanner}, which includes synthetic maps and diverse queries. Second, to demonstrate real-world applicability, we conducted a separate case study using OpenStreetMap~\cite{openstreetmap} data. The overall evaluation framework focuses on fine-grained route optimization and text-to-route bridging, using GPT-4o as the core agent LLM. Full details on the data construction process are provided in Appendix~\ref{app:map_construction}.

\underline{Map Environments.}
We construct synthetic yet realistic map environments that reflect diverse urban topologies, including dense city grids, suburban tree-structured networks, scenic tourist areas with loop routes, and mixed urban–rural transitions. Each map is a $50 \times 30$ grid with 8-directional connectivity, enriched with POIs (e.g., restaurants, coffee shops, gym, and park) and edge attributes such as distance, traffic, toll status, and scenic value. This controlled setup ensures reproducibility while providing sufficient diversity for evaluation.
\begin{table}[t]
\centering
\footnotesize
\caption{Parsing Analysis. Best results in \textbf{bold}.\\
\footnotesize Struct = schema valid rate, POI = POI\_F1, Const = Constraint\_F1, Pref = Preference\_F1}
\label{tab:parsing_results}
\begin{tabular}{l|ccccc}
\toprule
Method  & POI & Const & Pref & Struct &F1 \\
\midrule
Direct     & 0.905  & 1 & 0.738 & 1 & 0.880 \\
CoT        &   0.905 & 1  & 0.377 & 1 &  0.761\\
RouteLLM    &  \textbf{0.915} & \textbf{1}  & \textbf{0.788} & \textbf{1} &  \textbf{0.901}\\
\bottomrule
\end{tabular}
\end{table}
 
\underline{Query Generation.}
To evaluate the system across varying levels of complexity, we generated 100 natural language routing queries with GPT-4o. These queries were stratified by difficulty into three tiers: Simple (40\%), Medium (40\%), and Hard (20\%). The difficulty level is defined by the complexity of the request, which includes the number of Points of Interest (POIs) to visit—up to 2 for Simple, 3 for Medium, and 5 for Hard—and between 2 to 5 requirements of basic to mixed types. Each query is annotated with a gold-standard parse that includes (1) required POIs and attributes, (2) path constraints and weighted preferences, and (3) special requirements such as time-sensitive or weather-dependent conditions. We ensured that every generated query has at least one feasible solution.

\begin{figure*}[!t]
  \centering
  \begin{subfigure}{0.24\textwidth}
    \includegraphics[width=\linewidth]{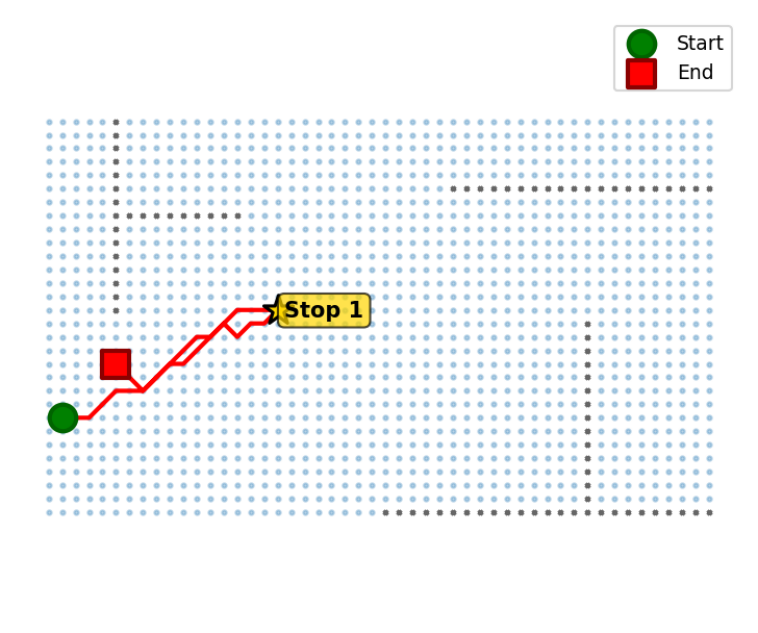}
    \caption{SP1}
    \label{fig:case-SP1}
  \end{subfigure}\hfill
  \begin{subfigure}{0.24\textwidth}
    \includegraphics[width=\linewidth]{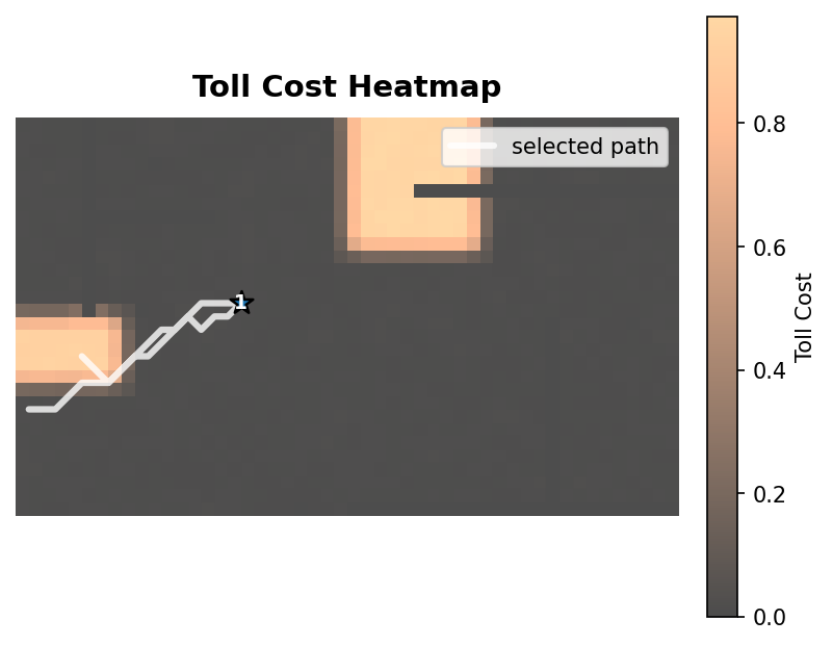}
    \caption{SP2}
    \label{fig:case-SP2}
  \end{subfigure}\hfill
  \begin{subfigure}{0.24\textwidth}
    \includegraphics[width=\linewidth]{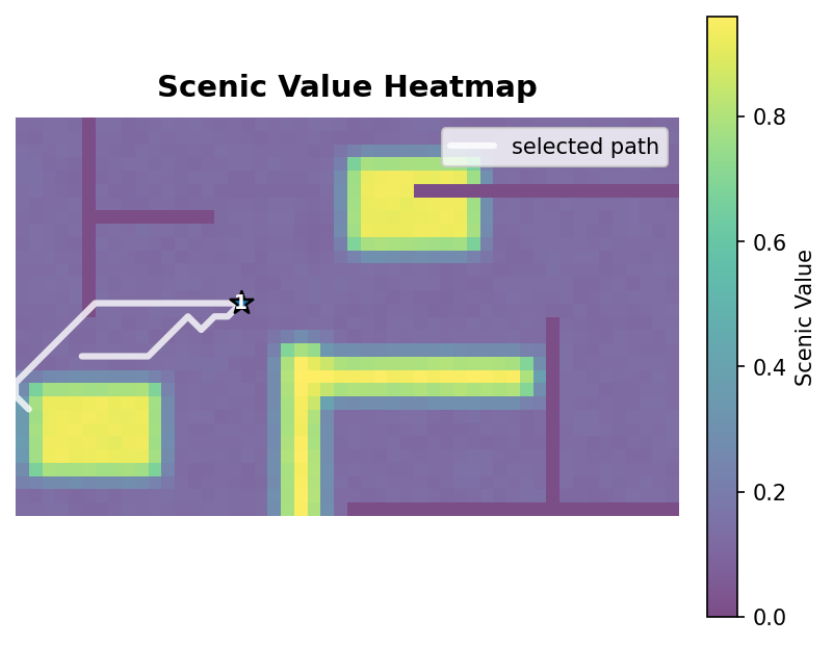}
    \caption{SP3}
    \label{fig:case-SP3}
  \end{subfigure}\hfill
    \begin{subfigure}{0.24\textwidth}
    \includegraphics[width=\linewidth]{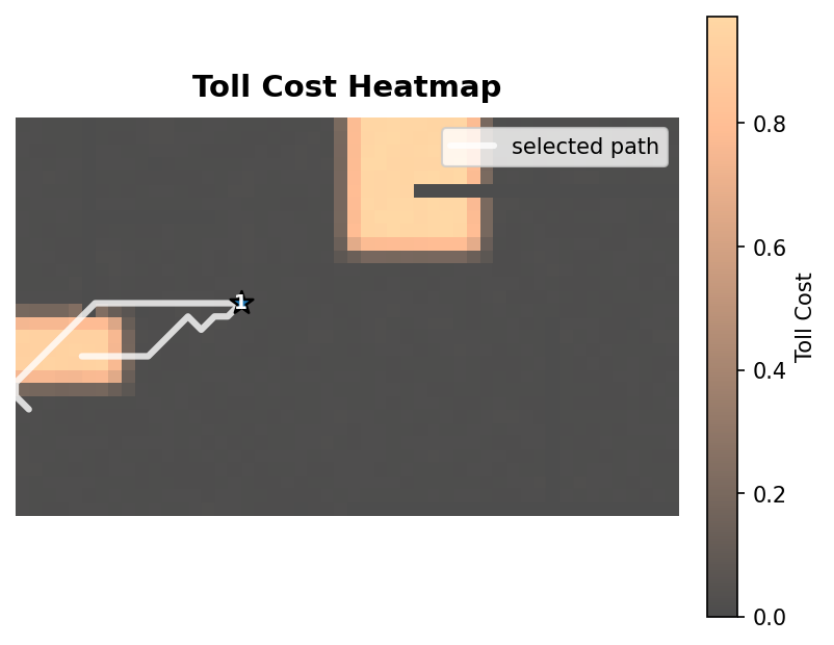}
    \caption{SP4}
    \label{fig:case-SP4}
  \end{subfigure}
  \caption{SP1: Path generated from the original; SP2: Same path overlaid on the toll cost heatmap;  SP3: Path generated after modifying the query to prioritize scenic value, shown on the scenic value heatmap;
  SP4: Scenic-prioritized path overlaid on the toll cost heatmap.}
  \label{fig:case-H002}
\end{figure*}
\begin{table*}[t]
\centering
\caption{Comparison of different scenarios on multiple cost dimensions. Lower is better.}
\begin{tabular}{lcccccc}
\toprule
\textbf{Scenario} & \textbf{cost\_danger} & \textbf{cost\_dist} & \textbf{cost\_energy} & \textbf{cost\_scenic} & \textbf{cost\_slope} & \textbf{cost\_toll} \\
\midrule
A\_Baseline      & 7.54 & 35.46 & 3.45 & 9.42 & 3.59 & 1.70 \\
B\_More\_Scenic  & 7.86 & 36.38 & 3.90 & 4.22 & 3.92 & 7.25 \\
C\_More\_Safe    & 7.54 & 35.46 & 3.45 & 9.42 & 3.59 & 1.70 \\
D\_Shortest\_Path & 7.41 & 32.97 & 3.54 & 8.59 & 3.63 & 2.60 \\
\bottomrule
\end{tabular}
\label{tab:scenario_costs}
\end{table*}

\begin{figure*}[!t]
  \centering
  \begin{subfigure}{0.32\textwidth}
    \includegraphics[width=\linewidth]{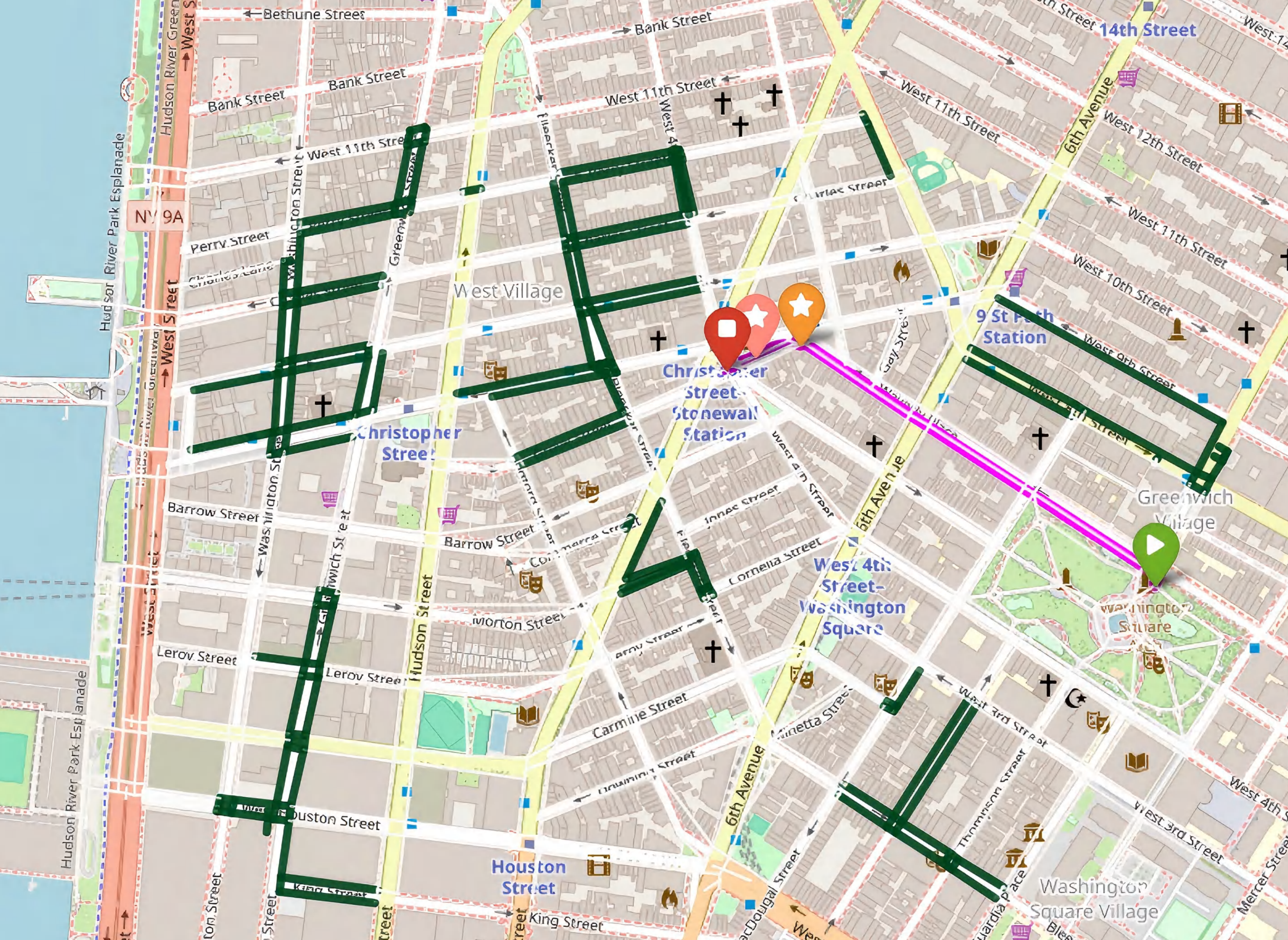}
    \caption{R1}
    \label{fig:case-r1}
  \end{subfigure}\hfill
  \begin{subfigure}{0.32\textwidth}
    \includegraphics[width=\linewidth]{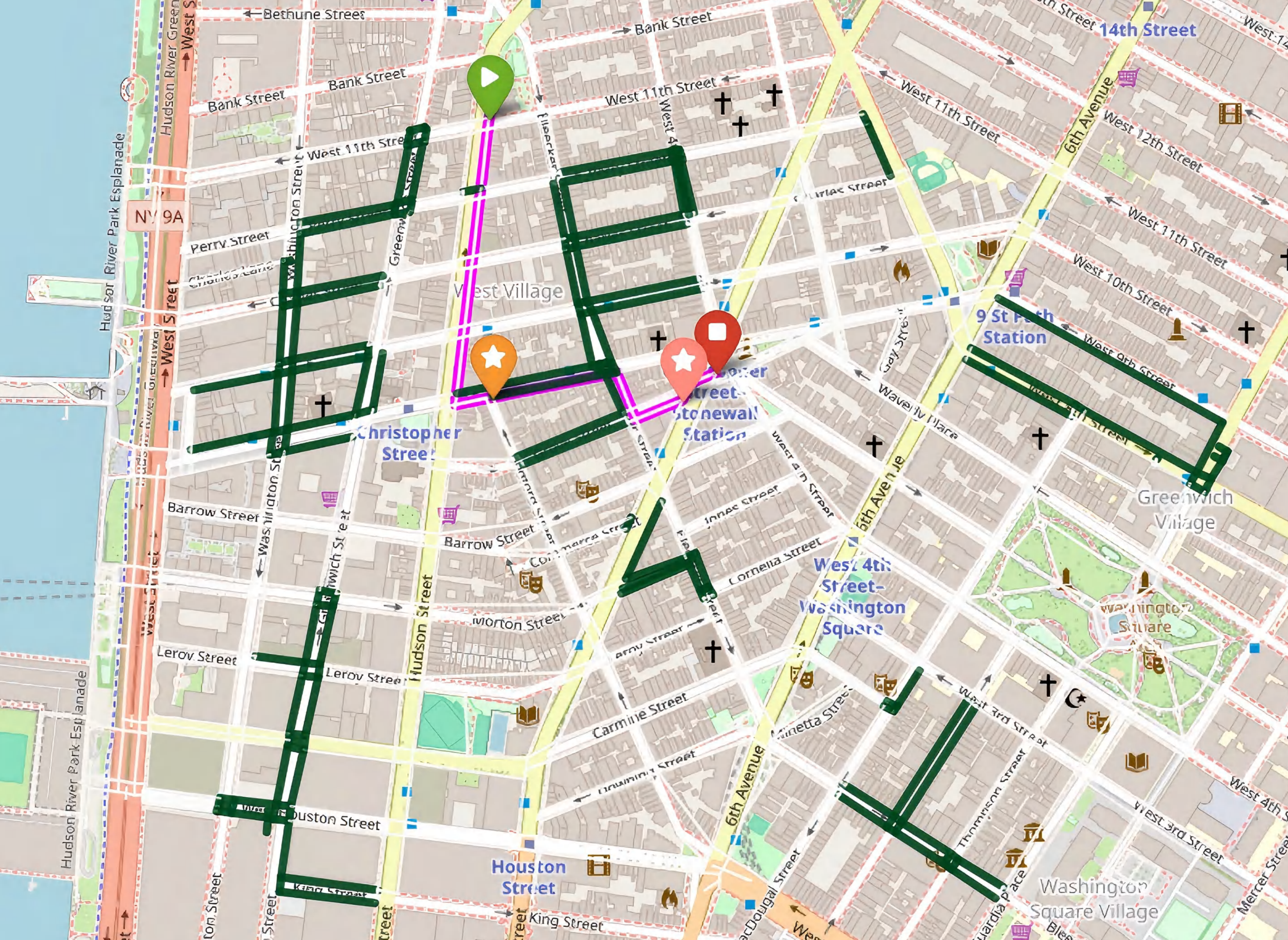}
    \caption{R2}
    \label{fig:case-r2}
  \end{subfigure}\hfill
  \begin{subfigure}{0.32\textwidth}
    \includegraphics[width=\linewidth]{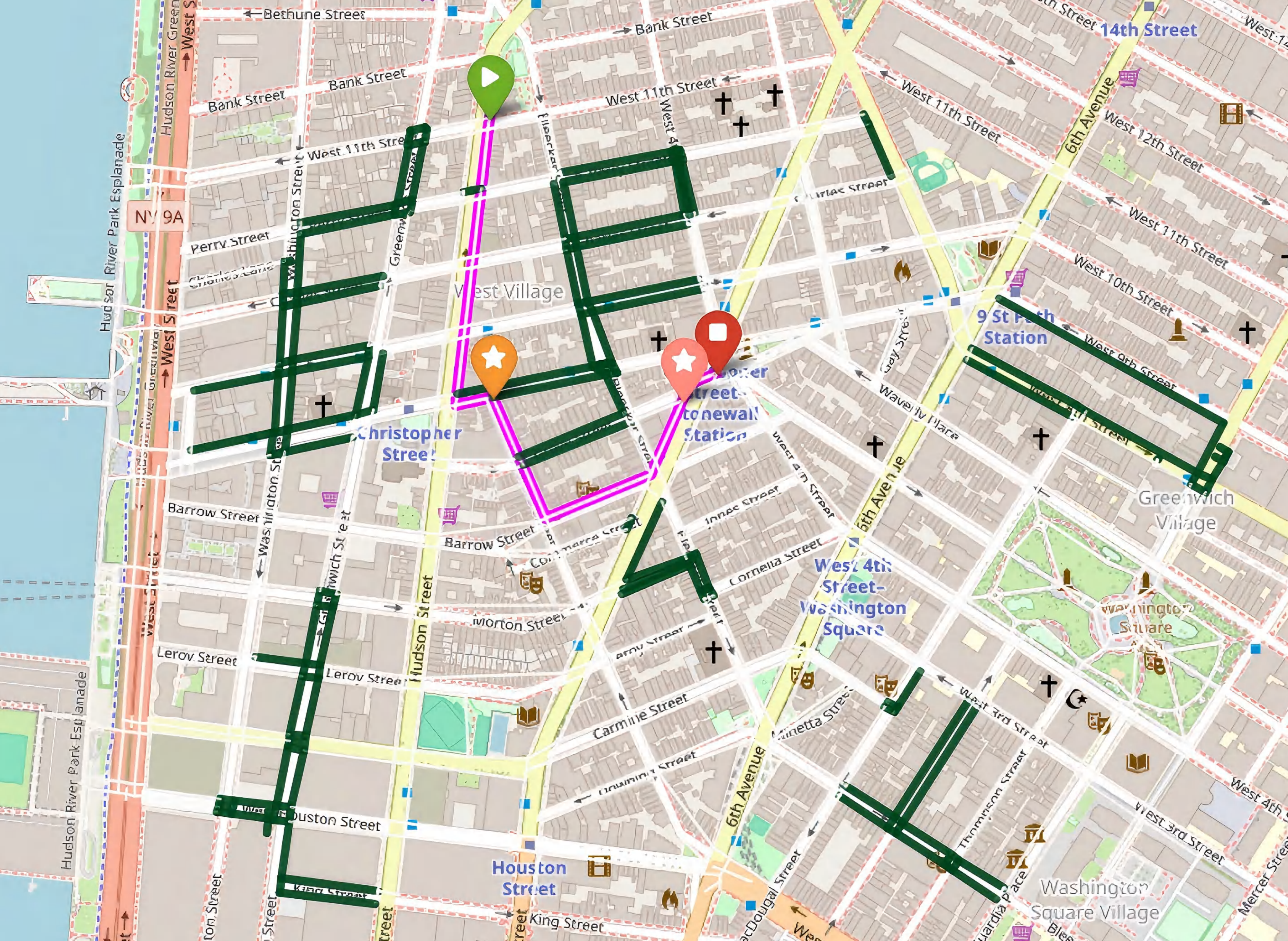}
    \caption{R3}
    \label{fig:case-r3}
  \end{subfigure}
  \caption{Case study on NYC Greenwich Village. 
  Pink lines show generated routes, bubbles mark POIs and start/end points, green lines indicate high-scenic cost streets. (a) R1: Route after changing start point; (b) R2: Route prioritizing shorter paths; (c) R3: Route prioritizing scenic views.}
  \label{fig:case-nyc}
\end{figure*}

\subsection{Parsing Reliability Analysis}
\textit{This experiment analyzes the parsing reliability of the agent}.
To assess our parsing approach, we compare \model's structured decomposition against two LLM prompting techniques: (1) \textbf{Direct}: which uses vanilla prompting without structured guidance, (2) \textbf{CoT}~\cite{wei2022chain}: which employs chain-of-thought reasoning. Our methods, (3) \textbf{RouteLLM}: utilize the [Object + Constraint] paradigm with CoT-guided reasoning and schema assurance via TrustCall.
For this evaluation, each query in our dataset is annotated with a gold-standard parse. Performance is measured by the F1 scores for three key extraction tasks—POI requirements (POI), path constraints (Const), and user preferences (Pref)—as well as the schema validity rate (Struct). The overall F1 score is the average of the three component extraction scores.
As shown in Table~\ref{tab:parsing_results}, all methods achieve perfect schema compliance (Struct = 1), ensuring structured outputs for downstream agents. 
However, baseline methods struggle with semantic extraction on diverse or ambiguous queries. Direct prompting achieves only 0.738 F1 on preference extraction, while CoT performs even worse at 0.377, suggesting that generic reasoning prompts may not adapt well to nuanced preference interpretation. Our RouteLLM parser addresses this through domain-specific contextual prompting, achieving 0.788 F1 on preferences and 0.901 overall F1.

\subsection{Route Adaptation under User Preferences}
\textit{This experiment demonstrates how changes in user preferences expressed in text are bridged to downstream route adaptations.}  
We evaluate the system's ability to adapt routes based on varying user preferences. Starting with a baseline query (Scenario A), we modify the request to prioritize different objectives: scenic routes (B), safety (C), and shortest distance (D). Figure~\ref{fig:case-H002} and Table~\ref{tab:scenario_costs} present the results.
\textit{Baseline vs. Scenic Preference:} The baseline route (SP2 in Figure~\ref{fig:case-H002}b) minimizes toll costs(1.70) but scores poorly on scenery (9.42). When the user prioritizes scenery (Scenario B), the system generates a new path (SP3 in Figure~\ref{fig:case-H002}c) that navigates through high-value scenic areas (yellow regions in the heatmap). This adaptation improves the scenic cost from 9.42 to 4.22, but increases toll costs to 7.25 (SP4 in Figure~\ref{fig:case-H002}d). This trade-off demonstrates the system's multi-objective balancing capability.
\textit{Other preference scenarios:} Scenario C (safer route) produces identical costs to the baseline, indicating the original route was already optimal for safety. Scenario D (shortest path) reduces distance cost from 35.46 to 32.97, with tolls increase to 2.60. These variations confirm the system responds appropriately to different preference weights.
The results demonstrate two critical capabilities: (1) accurate translation of natural language preferences into quantifiable route objectives, and (2) transparent trade-offs between competing criteria. This traceability enables users to understand why \model made specific routing decisions, supporting interpretability and iterative refinement.

\subsection{Case Study}
\textit{This experiment validates the framework on real-world street networks.}  
We deploy \model on New York City's Greenwich Village using real map and POI data from OSMnx~\cite{boeing2025modeling}. Figure~\ref{fig:case-nyc} shows routes generated from natural language queries (full queries in Appendix~\ref{app:NYC Queries}).
\textit{Dynamic query adaptation.} The framework responds to both structural and preference-based query changes. Figure~\ref{fig:case-r1} and Figure~\ref{fig:case-r2} demonstrate basic adaptability: when the user modifies the starting location, the system regenerates the route accordingly. This responsiveness to concrete query edits is essential for interactive route planning.
\textit{Preference-driven adaptation:} Figure~\ref{fig:case-nyc} demonstrates how preference changes affect routing decisions. When prioritizing shorter paths (R2), the system generates a direct route that passes through high-scenic (marked in green) cost streets. When emphasizing scenic views (R3), the route deviates to follow high-scenic streets  despite increased distance. This contrast confirms the framework translates natural language preferences into measurable route trade-offs.
The case study validates two aspects of real-world applicability. First, the system handles complex urban street networks with realistic POI distributions. Second, it adapts routes based on preference nuances expressed in natural language, not just binary constraints. These results suggest the framework generalizes beyond controlled environments to realistic deployment scenarios.

%% file: 6-Conclusion.tex
\section{Conclusion}
In this work, we introduced \model, a novel multi-agent framework that bridges natural language route requests with precise, constraint-aware route recommendations.
Through extensive experiments, we demonstrated its ability to reliably parse complex queries, adapt to user preferences, and generalize to real-world scenarios. Our results highlight the strong potential of combining LLM-driven reasoning with classical path optimization to deliver flexible yet robust route planning. Future work will explore scaling to larger, real-world datasets and extending personalization capabilities.

%% file: 7-Appendix.tex
\clearpage
\section{Appendix}

\subsection{Map Environment Construction}
\label{app:map_construction}
\paragraph{Grid Map Generation}
Our benchmark employs a set of synthetic yet realistic grid maps designed to test the framework under diverse conditions. Each map is defined as a $50 \times 30$ grid with 8-directional connectivity. This scale was chosen to be computationally manageable while remaining complex enough to simulate a variety of real-world urban topologies.
To facilitate multi-objective planning, each edge in the grid is assigned a vector of six cost attributes: distance, scenic value, energy consumption, danger level, slope, and toll cost. All attribute values are normalized to a range of [0,1], where a higher value uniformly represents a higher cost (e.g., lower scenic quality or greater danger).
To better imitate real-world scenarios, where features like scenic views or dangerous areas are often spatially clustered rather than uniformly random, we developed a two-step assignment process. First, all edges are initialized with a low base cost, drawn randomly from a narrow distribution around 0.1. Second, we designate several polygonal "zones" of varying shapes on the map for each attribute. Edges within these zones are assigned a high cost, drawn from a distribution around 0.9. This method ensures that features exhibit spatial coherence, creating more realistic and challenging scenarios for the routing agent.
Finally, to simulate urban obstacles, a small number of edges are designated as impassable. However, we ensure that the overall graph remains fully connected, guaranteeing that a feasible path exists between any two nodes.

\paragraph{Grid Map POI Generation}
To create a functional environment for routing queries, we populate the grid map with 50 Points of Interest (POIs). These are divided into four categories to simulate a variety of common destinations: 20 Restaurants, 15 Coffee Shops, 10 Gyms, and 5 Parks.
Each POI is defined by more than just its coordinates and category. To support the complex, multi-constraint queries that \model is designed to handle, each category is enriched with a set of specific attributes. For instance, as detailed in Table~\ref{tab:poi_schema}, 'Restaurants' include fields for cuisine, rating, and is\_vegetarian\_friendly, while 'Coffee Shops' have a boolean is\_work\_friendly tag to indicate the presence of WiFi and a quiet environment. After defining this schema, we utilized GPT-4o to populate the 50 POI entries with diverse and plausible values for each attribute, such as generating unique restaurant names, ratings, and specific opening hours.
Spatially, these 50 POIs are scattered across the grid in a roughly uniform random distribution, avoiding any specific clustering. This placement is a deliberate design choice to ensure the pathfinding environment is both challenging and realistic, creating a mix of short-distance (nearby POIs) and long-distance (far-apart POIs) planning scenarios.

\paragraph{Grid Map Query Generation}
The 100 natural language queries for our benchmark were generated with the assistance of GPT-4o. To achieve this, we provided the model with the context of our synthetic environment, including the map topology, POI locations, and the full schema of attributes. We then prompted GPT-4o to generate a diverse set of user routing queries that adhered to the difficulty stratification (40\% Simple, 40\% Medium, 20\% Hard) outlined in the Experimental Setup. Following this automated generation, each query underwent a manual curation process. During this step, we refined the wording for clarity and realism and, critically, verified that at least one feasible solution existed for each query on its corresponding map. Queries that were ambiguous or found to be unsolvable were discarded and regenerated.

\paragraph{NYC Greenwich Village Case Study Map Generation}
For the case study, we constructed a map environment based on a real-world location: New York City’s Greenwich Village. We used the OSMnx~\cite{boeing2025modeling} library to retrieve the pedestrian street network and relevant Points of Interest (POIs), such as restaurants and cafes, within a specific geographic bounding box(-74.0130, 40.7280, -73.9960, 40.7360).

A primary challenge when using real-world OSM data is its inherent complexity and incompleteness; many POIs lack the specific attributes (e.g., rating, average\_cost) required for our multi-constraint queries. To create a usable and realistic dataset, we implemented a multi-step data preparation pipeline:
\textit{Standardization:} We first normalized raw data fields that were inconsistent. For example, diverse cuisine tags (e.g., 'pizza', 'sushi') were mapped to a standard set (e.g., 'Italian', 'Japanese'), and varied opening\_hours strings were parsed into a consistent "HH:MM-HH:MM" format.
\textit{Attribute Enrichment:} To address missing data, we programmatically generated plausible values for critical attributes. For instance, missing rating and average\_cost values were simulated by drawing from normal distributions with category-specific parameters (e.g., cafes were assigned a higher average rating than fast-food restaurants). This ensures that every POI has a complete set of attributes for the agent to reason about.
\textit{Grid Abstraction:} Finally, to make the data compatible with our framework's underlying representation, the real-world longitude and latitude coordinates of each POI were mapped onto our $50 \times 30$ grid.
This rigorous process resulted in a case study environment that is both grounded in a real-world location and fully compatible with our system's data requirements, enabling a meaningful evaluation of its applicability.

\subsection{Queries for Case Study}
\label{app:NYC Queries}
The case study utilized a set of carefully designed natural language queries to test RouteLLM's adaptability in a real-world scenario. The queries were structured to evaluate the system's response to both concrete parameter changes and nuanced preference shifts. The key queries used to generate the results in Figure~\ref{fig:case-nyc} are summarized below:

\underline{R1 Query:} Hi there, can you plan a walking route for me this evening? I'm starting at the Washington Square Arch and need to end up at the Christopher Street-Sheridan Square subway station. First, I'd like to find a nice Italian restaurant for dinner. It should be pretty good-at least 4.0 stars-and have some good vegetarian options. My budget is around \$80. After dinner, I'd like to grab a drink at either a quiet bar or a nice cafe before I head to the subway. For the route itself, I'm not in a huge rush, so I'd really prefer to walk along the more scenic and well-lit (safer) streets through the Village. Thanks!"

\underline{R2 Query:} Hi there, can you plan a walking route for me this evening? I'm starting at the Chelsea Market and need to end up at the Christopher Street-Sheridan Square subway station. First, I'd like to find a nice Italian restaurant for dinner. It should be pretty good-at least 4.0 stars-and have some good vegetarian options. My budget is around \$80. After dinner, I'd like to grab a drink at either a quiet bar or a nice cafe before I head to the subway. For the route itself, I'm not in a huge rush, so I'd really prefer to walk as short as possible, tho I want well-lit (safer) streets through the Village. Thanks!"

\underline{R3 Query:} "Hi there, can you plan a walking route for me this evening? I'm starting at the Chelsea Market and need to end up at the Christopher Street-Sheridan Square subway station. First, I'd like to find a nice Italian restaurant for dinner. It should be pretty good-sat least 4.0 stars-and have some good vegetarian options. My budget is around \$80. After dinner, I'd like to grab a drink at either a quiet bar or a nice cafe before I head to the subway. For the route itself, I'm not in a huge rush, so I'd really prefer to walk along the more scenic and well-lit (safer) streets through the Village. Thanks!"

\subsection{No-Detour Spatial Filter}
To improve practical application, we designed an optional POI filtering strategy to avoid unnecessary detours. Although some POIs might fit the user's requirements, they can introduce inefficient travel. Crucially, to ensure the final route is logical and efficient, the agent employs a no-detour filtering mechanism. This spatial heuristic evaluates the arrangement of candidate POIs and eliminates orderings that create an inefficient path—for example, preventing a sequence like A→C→B when the optimal geographic progression is A→B→C. We find this specialized process of structured querying and spatial filtering is more reliable than asking a generalized LLM to reason about locations directly on a map.

\newpage
\subsection{Prompts for Parsing Reliability Analysis}

\begin{instructionbox}[red][Direct Prompting]
You are a travel planning assistant. Your task is to extract the user's requirements plan into a single, valid JSON object that strictly follows the provided schema format.

[The Common Definitions for POIs, Preferences, and JSON Structure were inserted here.]

Do not add explanations or markdown.
\end{instructionbox}

\vspace{32pt}
\begin{instructionbox}[blue][Chain-of-Thought Prompting]
You are an expert JSON parsing assistant. First, reason through the user's request by following these steps in your head:

1. Identify POI requirements using [Object, Constraints] paradigm.

2. Extract route preferences.

3. Note any hard limits.

[The Common Definitions for POIs, Preferences, and JSON Structure were inserted here.]

[Two detailed examples of user requests and the corresponding correct JSON output were provided here.]

After your reasoning is complete, output ONLY the raw JSON object. Do not include your reasoning steps.
Do not add explanations or markdown.

\end{instructionbox}

\newpage
\begin{instructionbox}[green][\model Prompting]
You are a highly precise and structured data extraction agent. Your sole purpose is to convert a user's travel request into a single, valid JSON object.

$<CONTEXT>$

[Context description was provided here.]

$</CONTEXT>$
\\[0.3cm]
$<SCHEMA\_AND\_RULES>$

[The target JSON schema and formatting rules were provided here.]

$</SCHEMA\_AND\_RULES>$
\\[0.3cm]
$<INSTRUCTIONS>$

Step 1: Parse Route Preferences (soft\_prefs)

[Rules for assigning preference weights were provided here.]

Step 2: Parse POI Stops (poi\_stops)

[Detailed instructions for handling sequential (AND) vs. optional (OR) logic were provided here.]

Step 3: Apply POI Filters

$<POI\_SPECIFICATION>$

[The detailed POI attribute schema was provided here.]

$</POI\_SPECIFICATION>$

$</INSTRUCTIONS>$
\\[0.3cm]
$<FEW\_SHOT\_EXAMPLES>$

[Two detailed examples of user requests and the corresponding correct JSON output were provided here.]

$</FEW\_SHOT\_EXAMPLES>$

\end{instructionbox}

\clearpage
\begin{table}[htbp]
\centering
\caption{Schema of POI Attributes by Category}
\label{tab:poi_schema}
\begin{tabularx}{\textwidth}{@{} l l X @{}}
\toprule
\textbf{Category} & \textbf{Field} & \textbf{Description / Possible Values} \\
\midrule
\multirow{7}{*}{Restaurant} & \texttt{coordinate (x,y)} & Grid coordinate of the POI (x $\in$ [0,49], y $\in$ [0,29]) \\
 & \texttt{category} & "restaurant" \\
 & \texttt{cuisine} & Cuisine type -- one of ["Chinese", "American", "Italian", "Mexican", "Indian", "Mediterranean", "French"] \\
 & \texttt{rating} & Customer rating, float between 0.0 and 5.0 (one decimal place) \\
 & \texttt{average\_cost} & Approx. average cost for a meal (integer, in currency units) \\
 & \texttt{is\_vegetarian\_friendly} & Boolean indicating if vegetarian options are available (\texttt{true / false}) \\
 & \texttt{opening\_hours} & Opening hours as a time range string in "HH:MM-HH:MM" (24-hour format) \\
\midrule
\multirow{5}{*}{Coffee Shop} & \texttt{coordinate (x,y)} & Grid coordinate of the POI (x $\in$ [0,49], y $\in$ [0,29]) \\
 & \texttt{category} & "coffee\_shop" \\
 & \texttt{is\_work\_friendly} & Boolean indicating if it's work-friendly (e.g., has WiFi, quiet) \\
 & \texttt{average\_cost} & Typical cost of a coffee/snack (integer in currency units) \\
 & \texttt{rating} & Customer rating, float between 0.0 and 5.0 (one decimal place) \\
  & \texttt{opening\_hours} & Opening hours as a time range string in "HH:MM-HH:MM" (24-hour format) \\
\midrule
\multirow{6}{*}{Gym} & \texttt{coordinate (x,y)} & Grid coordinate of the POI (x $\in$ [0,49], y $\in$ [0,29]) \\
 & \texttt{category} & "gym" \\
 & \texttt{rating} & Customer rating, float between 0.0 and 5.0 (one decimal place) \\
 & \texttt{average\_cost} & Cost of membership or drop-in (integer in currency units) \\
 & \texttt{has\_swimming\_pool} & Boolean indicating if a swimming pool is available (\texttt{true / false}) \\
 & \texttt{opening\_hours} & Operating hours, format "HH:MM-HH:MM" (e.g., "06:00-22:00") \\
\midrule
\multirow{5}{*}{Park} & \texttt{coordinate (x,y)} & Grid coordinate of the POI (x $\in$ [0,49], y $\in$ [0,29]) \\
 & \texttt{category} & "park" \\
 & \texttt{has\_entry\_fee} & Boolean indicating if the park charges an entry fee \\
 & \texttt{rating} & Visitor rating, float between 0.0 and 5.0 (one decimal place) \\
 & \texttt{opening\_hours} & Open hours, format "HH:MM-HH:MM" (e.g., "05:30-19:30") \\
\bottomrule
\end{tabularx}
\end{table}